\pdfoutput=1

\documentclass[11pt]{article}

\usepackage[final]{acl}

\usepackage{times}
\usepackage{latexsym}

\usepackage[T1]{fontenc}

\usepackage[utf8]{inputenc}

\usepackage{microtype}

\usepackage{inconsolata}
\usepackage{lscape}
\usepackage{subfig}
\usepackage{graphicx}
\usepackage{hyperref}
\usepackage{booktabs}
\usepackage{float}
\usepackage{multirow}

\title{Modeling Orthographic Variation in Occitan's Dialects}

\author{Zachary William Hopton \\
  Language and Space Lab\\
  University of Zurich \\
  \texttt{zacharywilliam.hopton@uzh.ch} \\\And
  No{\"e}mi Aepli\\
  Department of Computational Linguistics\\ University of Zurich \\
  \texttt{naepli@cl.uzh.ch} \\}

\begin{document}
\maketitle
\begin{abstract}
Effectively normalizing textual data poses a considerable challenge, especially for low-resource languages lacking standardized writing systems.  In this study, we fine-tuned a multilingual model with data from several Occitan dialects and conducted a series of experiments to assess the model's representations of these dialects. For evaluation purposes, we compiled a parallel lexicon encompassing four Occitan dialects.
Intrinsic evaluations of the model's embeddings revealed that surface similarity between the dialects strengthened representations. When the model was further fine-tuned for part-of-speech tagging and Universal Dependency parsing, its performance was robust to dialectical variation, even when trained solely on part-of-speech data from a single dialect. Our findings suggest that large multilingual models minimize the need for spelling normalization during pre-processing. 
\end{abstract}

\section{Introduction}

Traditionally, natural language processing pipelines have been designed to reduce noise in the data during pre-processing, either by removing it entirely (i.e., as one may do with URLs) or by normalizing noisy forms. Normalization can either improve users' understanding of a text or serve as a system-internal process that is meant to reduce noise, allowing a model to better learn from the vocabulary presented during training \citep{costa-bertaglia-volpe-nunes-2016-exploring}. 

However, with many of the recent successes in text normalization coming from neural networks, such as sequence-to-sequence (seq2seq) models that map orthographic variants to canonical forms, normalization can become computationally costly \citep{lusetti-etal-2018-encoder}. Furthermore, such supervised methods are generally impractical for low-resource languages, as these languages often lack labeled datasets with standardized word forms. In the case of non-standardized languages, there is also the issue of not having canonical forms to standardize to. Consequently, recent research has shifted its focus to determining the necessity of orthographic normalization and identifying how noisy data might prove advantageous \citep{srivastava-chiang-2023-bertwich,aepli-sennrich-2022-improving,al-sharou-etal-2021-towards}.

\begin{figure}
    \includegraphics[width=0.95\columnwidth]{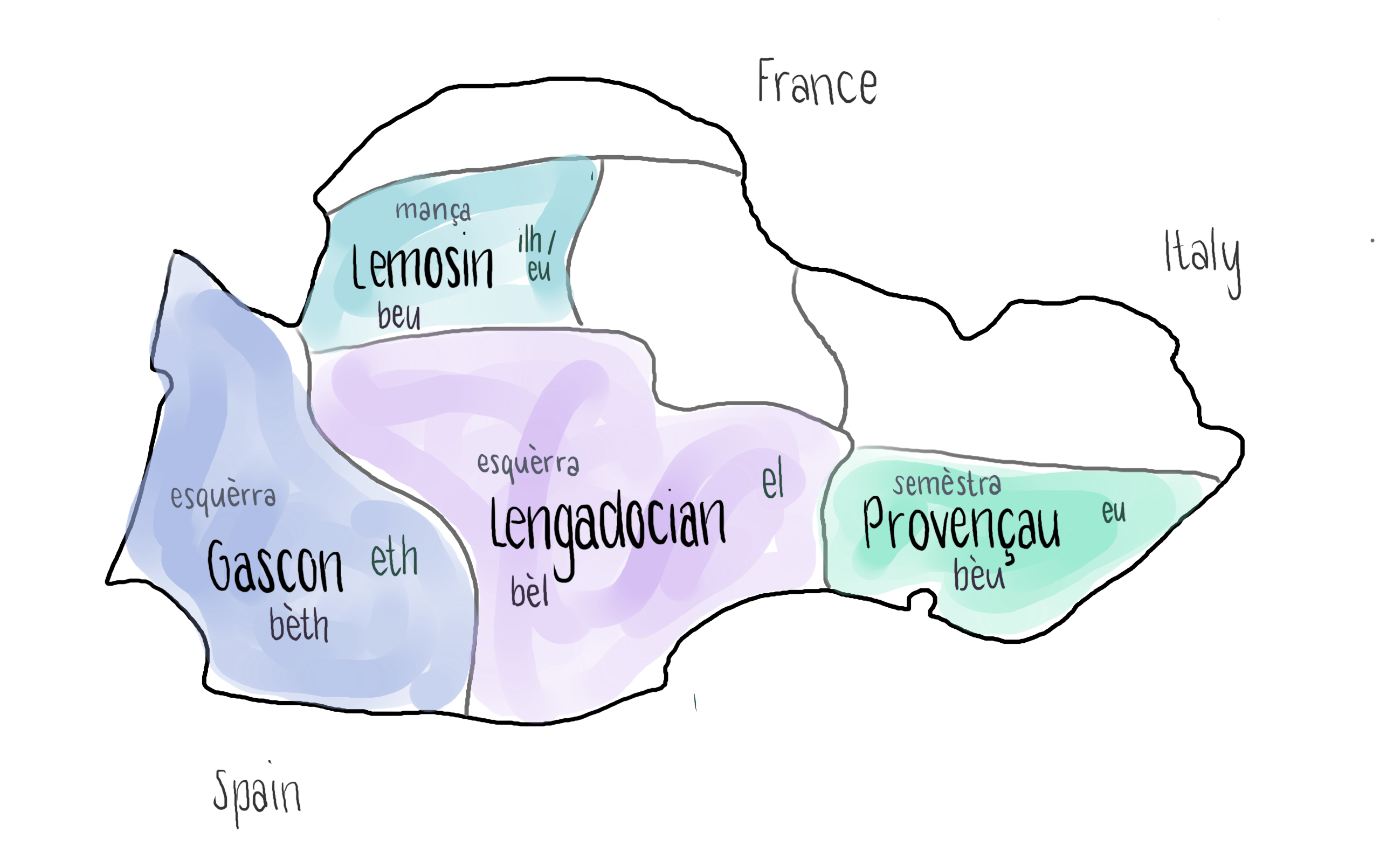}
    \caption{Dialect map of Occitan. The four dialects included in this study are highlighted, along with examples of lexical (i.e.,``mança'' and ``senèstra'') and spelling (i.e., ``bèu'' and ``beu'') variation between the dialects.}
    \label{fig:var_map}
\end{figure}

In the present study, we fine-tune a multilingual, large language model using data from various dialects of Occitan, a Western Romance language (see Figure \ref{fig:var_map}). We perform experiments to assess the model's ability to accurately represent the low-resource test language even without reducing the dialectal variation in the fine-tuning data, which manifests as both lexical and orthographic variation between dialects. Specifically, we carry out experiments on analogy computation and lexicon induction to test the fine-tuned model's intrinsic representations of Occitan's dialects, as well as Universal Dependency parsing and part-of-speech tagging experiments to study the usefulness of these fine-tuned representations in down-stream tasks. In doing so, we investigate the extent to which low-resource NLP systems that rely on transfer learning are robust to dialectal variation in fine-tuning data. This enables us to avoid costly and infeasible normalization during pre-processing.

\section{Linguistic Context}


Occitan is a Western Romance language spoken in southern France, the Val d'Aran in Spain, and Italy's Piedmont region. Occitan coexists in a diglossic relationship with French, Spanish, Catalan, and Italian and lacks official status outside of the Val d'Aran. The six dialects that are typically delineated are Auvernhàs, Gascon, Lemosin, Lengadocian, Provençau, and Vivaroaupenc \citep{bec1995langue}. Occitan is not standardized, and indeed, there is a great deal of geographical variation from many perspectives, including speakers' phonetic and lexical inventories, orthography, and syntax \citep{miletic-etal-2020-four}. At the level of phonetics, for example, Gascon stands out from other Occitan dialects and Western Romance languages with its use of the phone [h] \citep{corral-etal-2020-neural}. Occitan dialects also differ in their realization of /v/, with Lengadocian and Gascon speakers tending toward the phone [b] and Lemosin and Provençau speakers typically pronouncing [v] \citep{ocbook}.

As a whole, Occitan dialects share some morphological and syntactic features with each other, many of which are more similar to Catalan and Spanish than French. For example, unlike French, Occitan is a pro-drop language (i.e., subject pronouns are not necessary) and continues the use of the past preterit and imperfect subjunctive inflections outside of writing \citep{bernhard2021collecting}. Patterns of verbal inflection vary across the dialects, as do the augmentative and diminutive suffixes \citep{miletic-etal-2020-building, ocbook}. At the syntactic level, Gascon stands out with the use of enunciative particles to mark clauses' types \citep{morin2008syntax, vergez-couret-urieli-2014-pos}. 

A classical orthography inherited from medieval literature dominates Occitan writing in most dialects, though in Provençau, the ``Mistralian norm'' is more commonly used \citep{blanchet2004provenccal}. Some local writing systems have been standardized for purposes such as teaching \citep{bernhard2021collecting}. However, individuals vary in their conventions, often in ways heavily influenced by French orthography. Figure \ref{fig:var_map} highlights some of the variations in Occitan dialects' lexicons and spelling conventions. Besides the lack of a single orthographic convention, phonetic differences between the dialects lead to different spellings for words with the same meaning. For instance, in Provençau the word for \textit{``bedroom''} tends to be written as \textit{``cambra,''} and the initial /c/ is pronounced as velar [k] \citep{ocbook}. However, there is generally a palatalization of the consonant [k] among Occitan's Northern dialects such as Lemosin \citep{buckley2009phonetics}. This is often reflected in written forms, such as the Lemosin word for \textit{``bedroom,''} \textit{``chambra.''} 

Beyond just differences in spelling, dialects of Occitan vary at the lexical level. For instance, \textit{``achaptar''} (\textit{``to buy''}) is used in Lemosin while \textit{``crompar''} or \textit{``comprar''} are used in dialects to the south. Or, while speakers of Provençau tend to use the phrase \textit{``aver fam''} (\textit{``to be hungry''}), speakers in other dialects might say \textit{``aver talent.''}

These lexical and spelling variations between dialects of Occitan, along with a relative lack of data, pose challenges for NLP research. Nonetheless, there is a body of work on Occitan language technology, such as text-to-speech systems, part-of-speech taggers, universal dependency parsers, and lemmatizers \citep{corral-etal-2020-neural, vergez-couret-urieli-2014-pos, miletic-etal-2020-building,miletic-siewert-2023-lemmatization}. There is also a body of theoretical work about Occitan, such as experiments with continuous numerical representations of Occitan via cross-lingual word embeddings with related languages \citep{woller-etal-2021-neglect}. Most NLP research focusing on Occitan has been with four out of the six dialects that are generally delineated: Lengadocian, Lemosin, Provençau, and Gascon.

\section{Related Work}

At its simplest, normalization using rule-based word edits can collapse variants into standard forms \citep{reffle2011efficiently}. However, this requires language knowledge and becomes infeasible in cases of ambiguity. Thus, more context-sensitive approaches to text normalization, such as statistical string transduction and seq2seq neural networks, have been developed \citep{rios2014morphological, lusetti-etal-2018-encoder}. \citet{bawden-etal-2022-automatic} note that in both their statistical and neural machine translation approaches to normalizing Early Modern French, adding a rule-based post-processing step that constrains output to words in a contemporary French lexicon is particularly helpful. Moreover, \citet{lusetti-etal-2018-encoder} improved downstream machine translation scores for Swiss German following orthographic normalization with a character-level encoder-decoder model accompanied by a word-level language model. Recent work in normalizing both user-generated and multi-dialect data seems to confirm the effectiveness of working at the character and byte-level during normalization \citep{kuparinen-etal-2023-dialect,van-der-goot-etal-2021-multilexnorm}. However, framing normalization as a machine translation task requires large amounts of supervised data and is therefore not feasible in the case of many other low-resource languages, such as Occitan. 

Faced with an inability to remove orthographic noise to improve performance on downstream tasks, some have attempted to learn from the character level rather than the word or subword level. For instance, machine translation with character-level encoding can outperform subword-level encodings for morphologically rich languages, but requires deeper architectures, longer sequences, and—in models with no word or subword representations—becomes more difficult to interpret \citep{tang-etal-2020-understanding}. Despite these hurdles, another study has found that character representations result in better downstream machine translation performance than rule-based normalization of Swiss German training data \citep{honnet-etal-2018-machine}. Thus, character-level modeling seems to have potential as a substitute for spelling normalization during pre-processing. 

When modeling subword tokens, one way to better encode data with orthographic variation is by adding dropout to the byte-pair encoding (BPE) subwords \citep{sennrich-etal-2016-neural} during sub-tokenization. BPE-dropout \citep{provilkov-etal-2020-bpe} randomly removes a certain percentage of merges while applying BPE models to subtokenize a corpus. \citet{provilkov-etal-2020-bpe} found that applying BPE with dropout led to better machine translation of text with artificial misspellings. While exploring their models' embeddings, they find that subtokens in models trained with dropout tend to be similarly represented when they share sequences of characters. Thus, for training data with a non-standardized orthography, BPE dropout allows for more robust representations of spelling variants. 

Cross-lingual transfer learning is another potential means of alleviating the sparsity of data induced by orthographic variation. Recent work has shown that for Occitan, the quality of word embeddings can indeed be improved upon if jointly trained with more data from related languages, such as Catalan and Spanish \citep{woller-etal-2021-neglect}. Other work has sought to find normalization strategies for low-resource languages by pre-training classifiers with source languages different than the intended target language \citep{van2021cl}. This resulted in improvements over the baseline used, though interestingly, language pairs that performed the best were not always related languages. In recent work by \citet{aepli-sennrich-2022-improving}, the authors show that besides the relatedness of two languages, their surface similarity is also a critical factor for effective transfer learning. They find that by augmenting encoders' pre-training languages with random character noise, the models become more robust to spelling variation, and transfer learning is more effective based on performance on downstream tasks in the fine-tuning languages. The importance of surface similarity for transfer learning has also been highlighted in machine translation, where romanization of non-Latin scripts has been shown to improve the effectiveness of transfer learning when the pre-training and fine-tuning languages are related \citep{amrhein-sennrich-2020-romanization}. 

\section{Method}

\subsection{Creating a Dataset}

In order to conduct a controlled evaluation of our model after fine-tuning with multi-dialect Occitan data, we compile a parallel dataset comprising words from four Occitan dialects. 
This dataset allows us to explicitly compare the model's performance on the same content for each of the dialects. 
For the compilation of this dataset, we took inspiration from vocabulary themes in the book

\textit{Òc-ben! Première année d’Occitan—Livre d’éleve} \citep{ocbook}. We included various functional and lexical words, alongside short multi-word expressions and conjugations of both regular and irregular verbs in several tenses. In cases where a dialect was found to have variant spellings of the same word, we created multiple entries for that word rather than arbitrarily choosing between spellings. That being said, given that there is not a standardized writing system for Occitan, the dataset by no means captures the full breadth of possibilities for spelling variations, nor does it include information about which variants are preferred or more frequently used by Occitan speakers.

Our final parallel lexicon contains more than 2,200 entries in the Lengadocian, Lemosin, Provençau, and Gascon dialects. The dataset will be made available for use in academic research. 

\begin{figure}[h]
    \includegraphics[width=0.95\columnwidth]{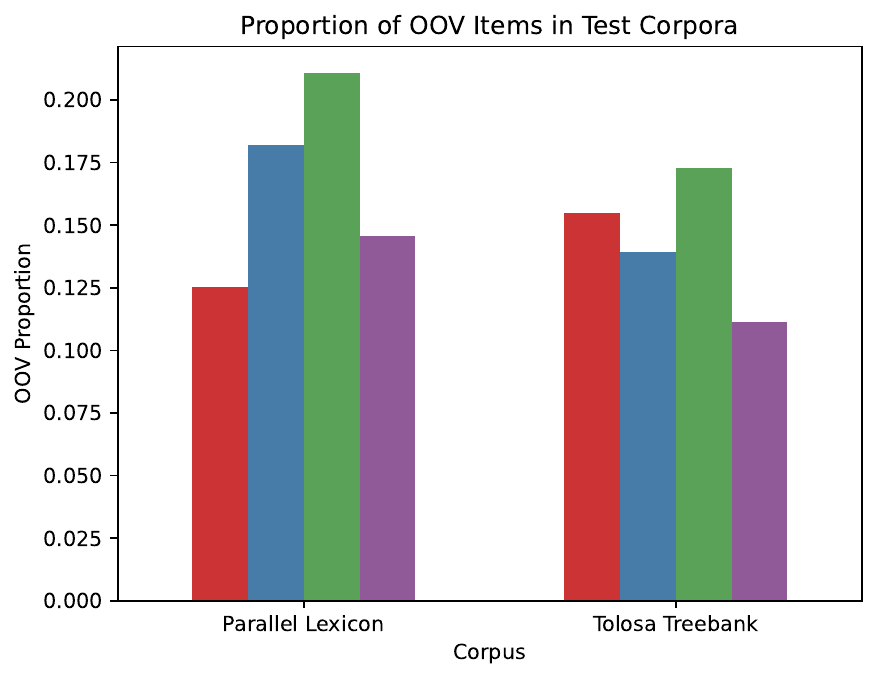}
    \caption{Proportion of vocabulary items in each evaluation corpus that did not appear in the fine-tuning dataset. Red: Lengadocian; Blue: Gascon; Green: Lemosin; Purple: Provençau.}
    \label{fig:oov}
\end{figure}

\subsection{Fine-Tuning mBERT}

We incorporate Occitan dialects into a language model by fine-tuning the multilingual BERT (mBERT\footnote{\url{https://huggingface.co/bert-base-multilingual-cased}}; \citet{devlin-etal-2019-bert}). mBERT has been pre-trained with both a masked language modeling objective and a sentence prediction objective on Wikipedia data from 104 languages. Our fine-tuning data comprises data from two sources: the OcWikiDisc corpus \citep{miletic-scherrer-2022-ocwikidisc}, which is compiled from Wikipedia discussion forums written in Occitan, and the WikiMatrix corpus \citep{schwenk-etal-2021-wikimatrix}, a corpus of parallel-sentence data mined from Wikipedia pages in 96 different languages. Annotation of a 100-sentence sample by the original authors of the OcWikiDisc corpus revealed that it contained several dialects of Occitan, though most of the sample came from the Lengadocian dialect. We specifically used the balanced OcWikiDisc corpus, which filtered the original dataset in a way designed to maximize the F1-score for language identification. From the WikiMatrix, we extracted the data from all language pairs containing Occitan and removed the parallel language data. Before combining with the OcWikiDisc data, we removed any duplicate lines from the dataset. 

The final combined fine-tuning corpus contained 386,552 lines (10,941,124 tokens) of data in Occitan. Ten percent of the training data was used as validation dataset during the fine-tuning process of three epochs. The ocwikidisc\_balanced corpus amounted to 756,922 (6.92\%) of the total tokens in the corpus. For details on the proportion of each test corpora's out-of vocabulary items with respect to the fine-tuning corpus, see Figure \ref{fig:oov}.

\section{Experiments}

\subsection{Analogy Representation}

\paragraph{Background} We conduct an intrinsic evaluation of the model's embedding space by assessing its representation of analogies. Using the parallel lexicon, we first created a dataset of analogies with approximately 35 data points per dialect. The data points were chosen to test linguistic relations similar to those presented by \citet{mikolov-etal-2013-linguistic}. Specifically, most of the relations were syntactic in nature, such as the relationship between infinitive verbs and conjugated forms, normal forms and diminutives or augmentatives, single and plural forms, and masculine and feminine forms. Ten of the data points for each dialect captured semantic relations, like antonym pairs and the relationship between capital city names and regions.

To test the model's representation of the analogies, we use two approaches described in \citet{levy-goldberg-2014-linguistic}. In the first approach, we use analogies in the form $a:b::x:y$ and seek the word $y$. Specifically, we search for the word $y$ in the given dialect's vocabulary whose embedding maximizes the following: 

\begin{equation}
    cos(y,b)-cos(y,a) + cos(y,x)
    \label{eq:levy-gold1}
\end{equation}

Similar to \citet{levy-goldberg-2014-linguistic}, we refer to Equation \ref{eq:levy-gold1} as ``3CosADD''.

As a second means of evaluating the analogies, we implement the ``3CosMUL'' metric from the same work by \citet{levy-goldberg-2014-linguistic}. The authors found that this multiplicative approach to combining the query vectors' meanings outperformed the additive approaches above. In this approach, we search for the word $y$ in a given dialect's vocabulary whose embedding maximizes the following:

\begin{equation}
    \frac{cos(y,b)cos(y,x)}{cos(y,a)+\epsilon}
    \label{eq:3cosmul}
\end{equation}

We set $\epsilon$ to 0.001—as in the original work—to avoid division by zero.

\paragraph{Results} When solving the analogies as set forward in Equation \ref{eq:levy-gold1}, the accuracies of both the base mBERT model and our fine-tuned model are poor across all four dialects (see Table \ref{tab:Analogies}). There is an overall increase in accuracy when using the 3CosMUL described in Equation \ref{eq:3cosmul}. The largest such improvement occurs for Gascon. 

\begin{table}
    \centering
    \small
    \begin{tabular}{r|cc}
         &3Cos-ADD & 3Cos-MUL\\ \midrule
         Gascon&  0.000 (-0.033)&  \textbf{0.133} (+0.066)\\
         Lengadocian& 0.097 (+0.065)&  \textbf{0.125} (+0.028)\\
         Lemosin& 0.069 (+0.035)&  \textbf{0.103} (+0.034)\\
         Provençau& 0.103 (+0.034)&  \textbf{0.138} (+0.035)\\
    \end{tabular}
    \caption{Fine-tuned model's accuracy in analogy computation, measured with two criteria. Values in parentheses represent the change in score from the baseline.}
    \label{tab:Analogies}
\end{table}

\begin{figure}
\small
\begin{tabular}{r|ll}
\multirow{2}{*}{\textbf{Semantic}}  & Fam : Minjar  & \textit{Hunger: Eat}                                              \\
                           & Set : Beure  & \textit{Thirst : Drink}                                            \\\midrule
\multirow{2}{*}{\textbf{Syntactic}} & Far : Fach  & \textit{Want (INF) : Want (PP)}                                      \\
                           & Voler : Volgut & \textit{Want (INF) : Want (PP)}

\end{tabular}
\caption{Examples of semantic and syntactic analogies from the Lemosin dataset with English translations in italics. INF: infinitive, PP: past participle.}
\label{fig:analogies_ex}
\end{figure}

\paragraph{Error Analysis} As a means of better understanding the results on our analogy dataset, we calculate the accuracy separately for the syntactic and semantic relations in our analogy dataset. See Figure \ref{fig:analogies_ex} for examples of each analogy type. The score on semantic analogies is 0.0 for all dialects, regardless of whether Equation \ref{eq:levy-gold1} or Equation \ref{eq:3cosmul} was used to solve the analogies. In Table \ref{tab:Analogies_synt}, we present the analogy scores for each dialect when only taking the syntactic relations into account. 

This disparity in performance for semantic and syntactic relations may be attributable to our evaluation approach. Indeed, \citet{levy-goldberg-2014-linguistic} note that there is an alternative formulation to the 3CosAdd approach called the ``PairDirection'' method. In their work on analyzing word embedding quality with analogy computations, \citet{mikolov-etal-2013-linguistic} used this PairDirection method for evaluating their semantic analogies while using a method that was algebraically equivalent to 3CosADD for syntactic relations. 

Beyond the method that we use to calculate the analogies, it would be interesting to experiment further with the specific pre-training tasks and architecture of our base model. For instance, the multi-network approach taken for pre-training sentence-BERT may offer a promising solution to the relatively poor representation of semantic relations in our fine-tuned model \citep{reimers-gurevych-2019-sentence}.

\begin{table}
    \centering
    \small
    \begin{tabular}{r|cc}
         & 3CosADD& 3CosMUL\\ \midrule
         Gascon&  0.0000& \textbf{0.2000}\\
         Lengadocian&  0.1429& \textbf{0.1818}\\
         Lemosin&  0.1053& \textbf{0.1579}\\
         Provençau&  0.1579& \textbf{0.2105}\\
    \end{tabular}
    \caption{Fine-tuned model's accuracy in analogy computation when only taking syntactic relations into account.}
    \label{tab:Analogies_synt}
\end{table}

\subsection{Lengadocian Lexicon Induction}

\paragraph{Background} As another means of evaluating the fine-tuned model's representations of Occitan's dialects, we conduct a lexicon induction task that assesses the similarity of parallel words across the dialects. Bilingual lexicon induction is a common use case for multilingual embeddings \citep{woller-etal-2021-neglect, mikolov2013exploiting}. Here, we aim to induce the Lengadocian lexicon using the other three dialects in our dataset.\footnote{We choose to induce the Lengadocian lexicon because it is likely the best-represented dialect in our training data, but the procedure could be repeated to induce any of the other dialects' lexicons.} For each word in Gascon, Provençau, and Lemosin, we find the Lengadocian word with the most similar embedding, again using cosine similarity. If the most similar word is an equivalent Lengadocian term for the other dialect's word, we score this as correct.

\paragraph{Results} Accuracy scores for the fine-tuned model's performance on the Lengadocian lexicon induction task can be found in Table \ref{tab:CrossDial}. Fine-tuning mBERT with multi-dialect Occitan data led to increases in performance on this task for all dialects. Interestingly, there is more disparity between dialects in this task compared to the 3CosMUL scoring of the analogy task. Whereas the fine-tuned model correctly selects the Lengadocian form of Provençau words in 40.9\% of the cases, performance for selecting the Lengadocian forms of Lemosin words is correct in just 29.1\% of cases.

\begin{table}
    \centering
    \small
    \begin{tabular}{r|c}
         &  Accuracy\\ \midrule
         Gascon&  0.322 (+0.067)\\
         Lemosin&  0.291 (+0.051)\\
         Provençau&  \textbf{0.409} (+0.109)\\
    \end{tabular}
    \caption{Fine-tuned model's accuracy in choosing a word's corresponding Lengadocian form (``Lengadocian Lexicon Induction''). Values in parentheses represent the change in score from the baseline.}
    \label{tab:CrossDial}
\end{table}

\paragraph{Error Analysis} To study the impact of dialectical variants' surface similarity on their representation, we stratify the results of the Lengadocian lexicon induction by the Levenshtein distance between the word in a given dialect and its counterpart in Lengadocian (see Table \ref{tab:CrossDialLev}). These results indicate that the further apart two word forms are, the less similarly our model represents them, even though they are semantically similar. 

\begin{table}
    \centering
    \small
    \begin{tabular}{r|ccc}
         &  Low& Med&High\\ \midrule
         Gascon&   \textbf{0.4982}& 0.2444&0.0917\\
         Lemosin&   \textbf{0.4172}& 0.2744&0.0756\\
         Provençau&   \textbf{0.5671}& 0.3766&0.1154\\
    \end{tabular}
    \caption{Fine-tuned model's accuracy in choosing a word's corresponding Lengadocian form (``Lengadocian Lexicon Induction''), stratified by orthographic distance between target Lengadocian word and corresponding dialect word; low: LevDist=1; Med: LevDist in range [2,3], High: LevDist > 3.}
    \label{tab:CrossDialLev}
\end{table}

This trend indicates that for cases where spelling differences are minimal, our fine-tuned model seems to model word pairs similarly. However, word pairs with low surface similarity are not represented well by the model. Thus, while our fine-tuned model may have learned to represent orthographical variation in Occitan well (i.e., variation of a few characters), it still struggles with dialectical variation at the level of whole lexical items. To some extent, Lemosin's relatively high proportion of OOV items with respect to the fine-tuning corpus (Figure \ref{fig:oov}) may explain the model's weaker performance in inducing the Lengadocian lexicon from Lemosin representations.

Table \ref{tab:LengLexErrors} contains examples of mistakes for the induction of the Lengadocian Lexicon from Provençau. Some mistakes are seemingly random, such as the Provençau word for ``January'' being closer to ``Give'' than the Lengadocian equivalent for January. However, an error such as the Provençau embedding for ``Neighborhood'' being most similar to the Lengadocian word ``Social'' shows some evidence of semantic consistency in our embeddings. Though this is just a single example, it shows that this method of evaluation for our embeddings is relatively strict. As emphasized by \citet{glavas2019properly}, if not done consistently and in the context of a more comprehensive analysis, bilingual lexicon induction is not necessarily an ideal evaluation of cross-lingual word embeddings.

\begin{table*}
\centering
\small
\begin{tabular}{c|c|c|c}
  & \textbf{True} & \textbf{Lengadocian Term} &  \textbf{Provençau-Target}\\
 \textbf{Provençau (\textit{EN})}& \textbf{Lengadocian} & \textbf{Selected in Evaluation (\textit{EN})} & \textbf{Levenshtein Distance} \\ \midrule
 Quartier (\textit{Neighbourhood}) & Quartièr & Social (\textit{Social})& 1 \\
 Fieu (\textit{String}) & Fial  & Lòc (\textit{Place}) & 2 \\
 Janvier (\textit{January}) & Genièr & Dar (\textit{Give (V.))} & 4 \\
\end{tabular}
\caption{Example errors from the Provençau–Lengadocian lexicon induction task. Column 2 contains the correct Lengadocian equivalent to the Provençau term; the incorrect Lengadocian term with the most similar embedding to the Provençau term is in column 3; English translations in parentheses.}
\label{tab:LengLexErrors}
\end{table*}

\subsection{Extrinsic Evaluation}
\label{sec:exeval}
\paragraph{Background} Using the Tolosa Treebank, a multi-dialect dataset \citep{miletic2020four}, we train task heads for part-of-speech tagging and Universal Dependency\footnote{\url{https://universaldependencies.org/}} parsing. The Tolosa Treebank contains texts from the Occitan dialects Lengadocian, Lemosin, Provençau, and Gascon. We experiment with two training setups: In the first, we use data from all four dialects in the train, validation,\footnote{Provençau and Lemosin are not included in the validation set due to the relatively small amount of data in these dialects.} and test sets. In the second setup, we attempt to mirror a more realistic scenario for low-resource languages where less annotated data is typically available. To do this, we train the task heads with only Lengadocian data. Lengadocian was chosen because the authors of the OcWikiDisc corpus believed this dialect to be the best represented in the corpus, and it has the most data in the Tolosa Treebank. We then test the PoS-tagging and dependency parsing abilities of the model on all four dialects with test sets from the Tolosa Treebank. As with the intrinsic evaluations, we report results for both the baseline mBERT model and our fine-tuned mBERT model. We use the MaChAmp framework \citep{van-der-goot-etal-2021-massive} for the multitask fine-tuning.

\paragraph{Results} Scores for the PoS taggers and dependency parsers are in Tables \ref{tab:PoS} and \ref{tab:UD_las}. In PoS tagging, accuracy is relatively high in both training scenarios. For Gascon, Lengadocian, and Provençau the fine-tuned model showed small improvements relative to the baseline mBERT, while the fine-tuned model performed slightly worse than the baseline for Lemosin. On the full test set with data from all four dialects, the model trained with PoS data from all four dialects reaches an accuracy of 94.1\%, outperforming the model trained on only Lengadocian data. PoS-tagging performance is best for Lengadocian data in both training setups, although only by a small margin.

As for UD parsing, Gascon had the highest labeled attachment scores in both training conditions, while performance was again the worst for Provençau. Similar to PoS tagging, scores for Lemosin dependency parsing decreased with fine-tuning. Across the dialects, there was a wider range of UD parsing scores compared to scores for PoS tagging, but performance was best on average in the condition where all four dialects were used during training. 
\begin{table}
    \centering
    \small
    \begin{tabular}{r|cc}
         &  All Dialects&Lengadocian\\ \midrule
         Gascon&   \textbf{0.941} (+0.005)&0.922 (+0.019)\\
         Lengadocian&   \textbf{0.946} (+0.008)&0.943 (+0.008)\\
 Lemosin& \textbf{0.934} (-0.002)&0.927 (-0.005)\\
         Provençau&   0.927 (+ 0.000)&\textbf{0.932} (+ 0.012)\\ \midrule
 Full Test Set& \textbf{0.941} (+0.005)&0.936 (+0.007)\\
    \end{tabular}
    \caption{PoS-tagging accuracy for the fine-tuned model; ``All Dialect'' PoS tagger used train and development data from all dialects. The ``Lengadocian'' PoS tagger was trained on only Lengadocian data. Values in parentheses represent the change in score from the baseline.}
    \label{tab:PoS}
\end{table}

\begin{table}
    \centering
    \small
    \begin{tabular}{r|cc}
         &  All Dialects&Lengadocian\\ \midrule
         Gascon&   \textbf{0.791} (+0.007)&0.755 (+0.035)\\
         Lengadocian&   \textbf{0.761} (+0.003)&0.742 (+0.008)\\
 Lemosin& \textbf{0.691} (-0.007)&0.679 (-0.012)\\
         Provençau&   \textbf{0.603} (- 0.005)& 0.579 (+0.003)\\ \midrule
 Full Test Set& \textbf{0.735} (+0.001)&0.714 (+0.008)\\
    \end{tabular}
    \caption{Labeled attachment scores for Universal Dependency parsing with the fine-tuned model. ``All Dialect'' parser used train and development data from all dialects. The ``Lengadocian'' parser was trained on only Lengadocian data. Values in parentheses represent the change in score from the baseline.}
    \label{tab:UD_las}
\end{table}

\begin{figure*}[h]
    \includegraphics[width=0.9\textwidth]{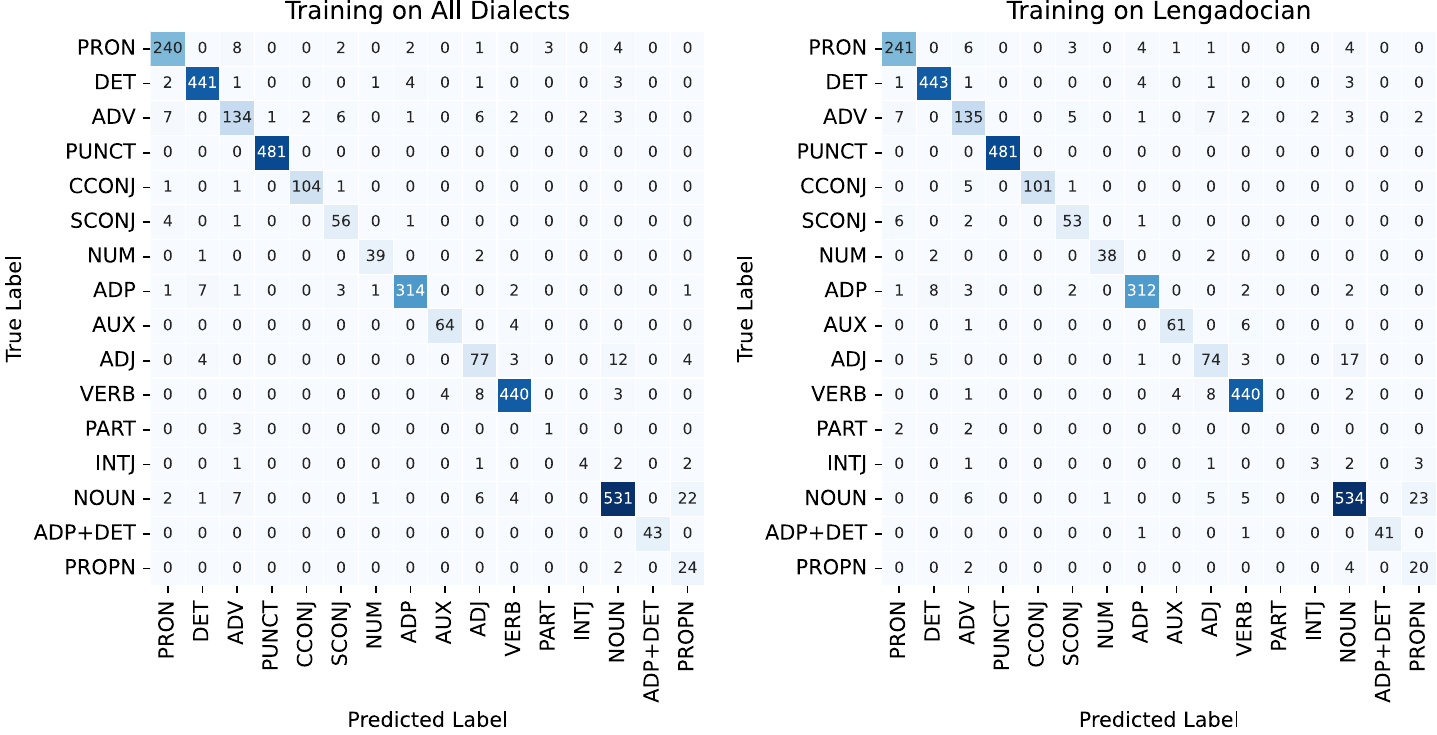}
    \caption{Confusion matrix for PoS taggers when trained on data from all dialects (left) and only Lengadocian (right).}
    \label{fig:cm}
\end{figure*}

\paragraph{Error Analysis} Despite observing similar results, we explore potential differences in the quality of the two part-of-speech taggers we trained. To do this, we visualize confusion matrices to illustrate how each tagger performed on the full test set, which included data from all four studied dialects (see Figure \ref{fig:cm}). 
The results highlight one key failure of our ``low-resource'' method where we use a PoS tagger that was trained only on Lengadocian data for inference on all dialects. Specifically, it indicates that the Lengadocian-trained tagger never correctly classifies particles in the evaluation set. This is an expected limit of zero-shot attempts to do PoS tagging with Gascon, as Gascon is the only dialect of Occitan that carries enunciative particles that mark clause type. This problem was also encountered in previous work on Occitan PoS tagging, where the authors proposed explicit rules for tagging the particles as a solution \citep{vergez-couret-urieli-2014-pos}. In the original dependency parsing experiments published with the Tolosa Treebank, the authors improve performance on Gascon evaluation data by including training data from Gascon in their model, an approach that also led to our highest-performing PoS tagger \citep{miletic-etal-2020-building}. 

In the same paper on the Tolosa Treebank, \citet{miletic2020four} note that performance for dependency parsing was consistently worst for Provençau. We obtained similar results for our dependency parsing experiments. As for our PoS taggers, when trained on all four dialects, performance was also worst on Provençau. Yet, the Provençau portion of the Tolosa Treebank has the lowest proportion of OOV items relative to the fine-tuning corpus, meaning we would generally expect its contents to be among the best represented of the dialects (Figure \ref{fig:oov}). To the contrary, performance on Provençau consistently outranks the other dialects for our intrinsic evaluations. We encourage further research on the specific morphosyntactic properties and orthographic tendencies of Provençau to clarify why—despite its relatively strong internal representation—it stands out as more difficult to tag in these tasks.

\section{Discussion}

Overall, our experiments on using non-standardized text data to fine-tune mBERT yielded mixed results. Fine-tuning mBERT with orthographically non-standard Occitan data led to little improvement in terms of performance on computing analogies, POS tagging, and dependency parsing. However, fine-tuning with the multi-dialect data consistently improved results in using one dialect's lexicon to induce the parallel lexicon of another dialect, Lengadocian. 

Taken together, these results provide support for the idea that including data with dialectical lexical variation and non-standardized orthography in fine-tuning data is not necessarily harmful to model performance. I.e., the fine-tuning we carried out with multiple dialects of Occitan did not deprecate mBERT's baseline performance on downstream tasks like part-of-speech tagging. As normalizing data during preprocessing can pose a substantial burden for low-resource NLP, our results are encouraging in that they suggest that in some contexts, including orthographically inconsistent data from multiple dialects will not harm the model.

Previous work on fine-tuning language models has led to various conclusions about the effect of different types of ``noise'' in the fine-tuning data. For instance, it has been shown that fine-tuning on English data with synthetic spelling errors can reduce BERT's performance on downstream sentiment analysis \citep{Kumar2020NoisyTD}. While our results do not seem to indicate a negative effect of including non-standardized data in the fine-tuning data, simply including small amounts of data from multiple dialects of Occitan was not enough to increase mBERT's performance on downstream tasks (PoS tagging and UD parsing) with the dialects. Furthermore, as shown in the Lengadocian lexicon induction, the model failed to capture the similarity of parallel lexical items that have low surface similarities (i.e., high Levenshtein distance).

Some research has shown that high surface similarity between pre-training and fine-tuning data can result in better performance on downstream tasks such as PoS tagging and machine translation \citep{aepli-sennrich-2022-improving, amrhein-sennrich-2020-romanization}. Bearing that in mind, our future efforts will look at whether increasing the surface similarity between the pre-training and Occitan fine-tuning data of our models will allow the model to better learn from the non-standardized data. Indeed, results from our Lengadocian induction task provide some further evidence that this may help, as the model represented parallel words from different dialects more similarly when their spelling was more similar. Increasing the surface similarity of the pre-training and fine-tuning data may involve using a model pre-trained only on languages that are more closely related to Occitan, such as Catalan, Spanish, and French. In the same vein as \citet{aepli-sennrich-2022-improving}, we may also explore injecting the pre-training data with noise in the form of characters or even replacing whole words with Occitan variants. Furthermore, \citet{Kumar2020NoisyTD} attribute BERT's detriment in performance when fine-tuned on noisy data to the model's tokenizer. Though out of the scope of the present work, we intend to focus future efforts on the overlap between the Occitan dialects' subtokens. Work on this matter may be particularly beneficial in understanding the results of our lexicon induction task. 

\section{Conclusion}

In this work, we experimented with the capacity of a pre-trained encoder to represent dialect variation—both lexical and orthographic—in a low-resource language. In doing so, we aimed to test the extent to which cross-lingual transfer learning allows for effectively representing Occitan's dialects. While the experiments yielded no clear evidence that dialect representation was improved after fine-tuning, we can still interpret our findings as an indicator that orthographic normalization may not be necessary when fine-tuning large, multilingual models. 

\section*{Limitations}

A potential limitation of this study is that we only worked with one base model, mBERT. It is possible that experimenting with other language models, i.e., models only on languages closely related to Occitan, would have yielded different results, while also telling us more about the relative usefulness of massively multilingual models for low-resource languages.

Another limitation is our inability to better characterize the pre-training and fine-tuning data in the experiment. Indeed, while the authors of the OcWikiDisc corpus performed manual evaluations to determine the dialect make-up of a small sample of their corpus, the total number of data points in each dialect in the OcWikiDisc is not known \citep{miletic-scherrer-2022-ocwikidisc}. Even less is known about the writing standard and dialect make-up of the WikiMatrix data which we also used for fine-tuning, meaning that overall, we cannot be sure that any variation between the dialects' results was not simply driven by a difference in the amount of data in each.

Further, our experimental setup is limited in that our UD parsers perform worse than the highest performing UD parsers trained on the Tolosa Treebank in \citet{miletic2020four}. The authors note that their worst LAS scores come from a model that was also trained on UD data from languages closely related to Occitan. Along the same lines, it may be that our use of a large, multilingual language model to carry out the UD parsing is limiting the utility of the relatively small amounts of dialect-specific Occitan UD data. 

\section*{Acknowledgements}

We would like to extend our gratitude to Aleksandra Mileti\'{c} for sharing the Tolosa Treebank, as well as to the anonymous reviewers for their insightful commentary.
This work was supported by the Swiss National Science Foundation (project no. 191934).

\bibliography{anthology,custom}



\end{document}